\documentclass[11pt]{article}

\usepackage{microtype}
\usepackage{graphicx,nicefrac}
\usepackage{subfig}
\usepackage{xcolor}
\usepackage{booktabs} %
\usepackage{algorithm}
\usepackage{algorithmic}
\usepackage{hyperref}

\usepackage{amsmath}
\usepackage{amssymb}
\usepackage{mathtools}
\usepackage{amsthm}

\usepackage[capitalize,noabbrev]{cleveref}

\theoremstyle{plain}

\theoremstyle{definition}

\theoremstyle{remark}

\usepackage[textsize=tiny]{todonotes}
\usepackage{placeins}

\advance \oddsidemargin by -0.9in
\newcommand{\myparskip}{3pt}
\parskip \myparskip

\newcommand\blfootnote[1]{%
  \begingroup
  \renewcommand\thefootnote{}\footnote{#1}%
  \addtocounter{footnote}{-1}%
  \endgroup
}

\begin{document}

\author{
Leo Betthauser$^*$
\and
Urszula Chajewska$^*$
\and
Maurice Diesendruck$^*$
\and
Rohith Pesala$^*$
}
\date{}


\title{Discovering Distribution Shifts using Latent Space Representations
\blfootnote{Equal contribution.\\Microsoft Corporation, 1 Microsoft Way, Redmond, WA 98052, USA.\\Code: https://github.com/microsoft/distribution-shift-latent-representations}
}
\maketitle


%
%
%
%

%
%
%
%

%
%
%





\vskip 0.3in

%

%
%
%
%
%
%

\begin{abstract}
Rapid progress in representation learning has led to a proliferation of embedding models, and to associated challenges of model selection and practical application. It is non-trivial to assess a model's generalizability to new, candidate datasets and failure to generalize may lead to poor performance on downstream tasks. Distribution shifts are one cause of reduced generalizability, and are often difficult to detect in practice. In this paper, we use the embedding space geometry to propose a non-parametric framework for detecting distribution shifts, and specify two tests. The first test detects shifts by establishing a robustness boundary, determined by an intelligible performance criterion, for comparing reference and candidate datasets. The second test detects shifts by featurizing and classifying multiple subsamples of two datasets as in-distribution and out-of-distribution. In evaluation, both tests detect model-impacting distribution shifts, in various shift scenarios, for both synthetic and real-world datasets.

\end{abstract}

\section{Introduction}
\label{sec:intro}
The field of Artificial Intelligence has made significant progress due to the introduction and refinement of representation learning models. 
These models produce vector representations of data (embeddings) that can be used in downstream tasks, such as similarity search, classification, regression, and image/language generation.

To perform well, models are often trained with large datasets and expensive computing resources that can be out of reach for many. As a result, pre-trained models are made available for public use, and practioners apply them in hopes that the knowledge encoded in the pre-trained model will transfer well to their application. Despite impressive results in technical reports, when applied to new data, these models produce embedding spaces that are minimally understood by their users, making it difficult to assess generalizability.

If practitioners have a downstream task in mind, they might assess the performance of that task; however, in cases of unsupervised or general-purpose models, embeddings are difficult to evaluate systematically. We would like to have an algorithm to determine a model's generalizability to new data, based only on the embeddings produced by that model.

Distribution shifts between original and candidate datasets are a common cause of reduced generalizability. That said,
the tolerance for a shift magnitude can vary from one domain to another.
A shift detection method should be flexible enough to accommodate the required shift sensitivity. Our proposed framework is flexible, practical in both runtime and hardware requirements, and easy to evaluate.

\begin{figure}[t]
    \centering
    \includegraphics[width=0.8\textwidth]{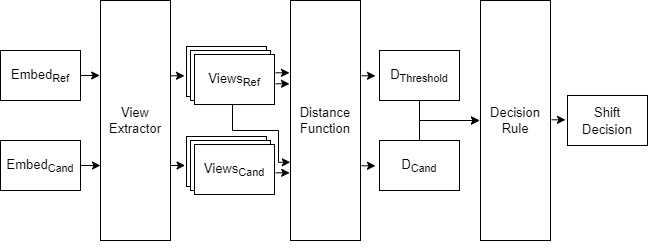}
    \caption{Distribution shift detection framework.}
    \label{fig:diagram}
\end{figure}

In the most general sense, the framework proposed (Fig.~\ref{fig:diagram}) begins with a single pre-trained model, which produces two sets of embeddings, one for a reference dataset, and another for a candidate dataset. From each embedding set, we extract multiple ``views" -- a view can be any randomized subsample and transform of the data; for example, a subsample with noise. A distance function then measures variation among reference views, and calls this the ``threshold"; and measures variation between reference and candidate views, and calls this the ``candidate distance". Finally, a decision rule compares the candidate distance to the threshold, to determine if a shift has occurred. 

More specifically, our method relies on finding a threshold distance $D^*$ between two distributions, for some suitable distance metric $D$, that determines the presence of a shift. We can find $D^*$ by perturbing reference embeddings in an increasing degree until the perturbed embeddings fail a performance test.
Alternatively we can also find $D^*$ by taking subsamples of the original embeddings to account for in-sample variation.
The choice of perturbation type, distance metric, performance test and threshold work with sensible defaults (provided below), and are tunable for specific applications.

In section \ref{sec:related} we present work most closely related to ours.  Section \ref{sec:approach} describes algorithms we propose for distribution shift detection and associated choices for metrics.  This approach is tested on both synthetic and real-world shifts.  The experimental design is described in section \ref{sec:setup} and results in section \ref{sec:results}.  We conclude in section \ref{sec:conclusion} with discussion and in section \ref{sec:future_work} with an outline of future work.

\section{Related Work}
\label{sec:related}
Distribution shifts are an inevitable problem in machine learning model deployment \cite{qui2009a}. Some well-known examples include subpopulation shifts where the test distribution is a subset of the training distribution, and domain generalization where it is the superset. 
While recent work has studied machine learning systems’ robustness to noise and transformation \cite{hendrycks2019a}, much less attention has been given to shifts observed in real-world deployments \cite{geirhos2020a}. A notable exception is the WILDS project 
\cite{koh-a} which published a set of benchmarks for studying distribution shifts in real-world settings. 
An effort to create a more specialized benchmark for EHR data is presented in \cite{DShift-EHR-Bench} and for Graph Neural Networks (GNNs) in \cite{DShift-Graph-Bench}.
A recent work \cite{DShift-BuresDiv} explores various metrics and approximations to compare two distributions, focusing on worst discrepancies between the two. \cite{DShift-OOD-Detection} propose a method for detection of out-of-distribution sample using geodesic (Fisher-Rao) distance.

There is even less work on discovering distribution shifts by analyzing embeddings.  \cite{vu-a} proposes a framework for evaluating different word embedding models for a given downstream task. In contrast, our approach is designed to work with arbitrary embeddings (including text, image, video, protein, etc.) and does not require access to a downstream task for evaluation.  \cite{DShift-AttribAlignment} propose a method for discovering a distribution shift in input data by learning a latent space of \textit{attributes}.  Their method could possibly be applied to embedding spaces as well.

The recent work of \cite{liu2020learning} and \cite{schrab2021mmd} in two-sample testing shares a similar goal of devising nonparametric tests for independence of distribution between two sets. Rather than design and optimize specific kernels for use with Maximum Mean Discrepancy-based tests, this work focuses on a class of tests amenable to many distance metrics and system performance criteria, that are also practical to implement and easy to evaluate.

\section{Approach}
\label{sec:approach}

We assume we are given two sets of embeddings, $X$ and $Y$ of dimension $d$, generated by the same model based on two datasets. Embeddings can represent any modality: word, document, image, video, graph, or an instance of tabular or time series data. Each representation consists of a vector $\vec{w}\in\mathbb{R}^d$. We will refer to $X$ as reference embeddings, $Y$ as candidate embeddings. 
Going forward, since all operations are on embeddings, the words ``embeddings" and ``datasets" are used interchangeably.

Our goal is to quantify the degree to which these two embedding sets indicate a distribution shift present in input data.

\subsection{Types of Distribution Shifts}

We use the term \textit{distribution shift} to refer to any situation where training and test data are generated by two different distributions.  Two most commonly encountered shifts are domain generalization and subpopulation shifts.  Domain generalization occurs when we train the model on data drawn from a subset of domains of interest.  For example, we may collect data on patients from a few hospitals and use it to make predictions for patients from the entire hospital network.  Or we could train a word embedding model on one type of medical text, e.g. PubMed articles, and apply it to different types of medical text, such as clinical trial descriptions or patient visit summaries. 

Subpopulation shift refers to the situation where the test dataset is a subset of the training dataset.  The goal is to do well on worst-case subpopulations, such as under-represented classes.

From a technical perspective, in both types of shifts, we may encounter a different distribution over classes (if classes are well defined and the dataset contains labels) and different density in some subspaces of the embedding space. We will explore these specific distributional differences in our experiments with synthetic shifts (section \ref{sec:synthetic}) and address domain generalization and subpopulation shifts generally in experiments on real-world data (section \ref{sec:real-world}).

\subsection{Distribution Shift Tests}
\label{sec:tests}

We evaluate $X$ and $Y$ for the presence of a distribution shift by computing a distance metric and comparing its value with a previously established threshold.  We propose two ways of computing an appropriate threshold:  using perturbations and using subsampling.

\textbf{Perturbation Shift Test.}
In this test (pseudocode presented as Algorithm~\ref{alg:perturbation}), we apply a perturbation to the reference embeddings.  We then compare the perturbed dataset to the original by using a simple performance test -- such as kNN recall -- with an appropriate performance threshold $c^*$.  As we increase the level of perturbation, our perturbed dataset will eventually fail the performance test.  We can apply a suitable distance metric to the original dataset and the perturbed dataset at the maximal level still meeting performance threshold, to arrive at distance threshold value $D^*$.  Since our perturbation methods are stochastic, to make the procedure more robust, we can generate multiple samples of the perturbed dataset at the desired perturbation level and aggregate values of $D$ for all samples.  Finally, we can compare $X$ and $Y$ using our distance metric $D$ to see if their distance is below threshold $D^*$. 

\begin{algorithm}[tb]
   \caption{Perturbation Shift Test}
   \label{alg:perturbation}
\begin{algorithmic}
   \STATE {\bfseries Input:} Embeddings $X$, $Y$, distance metric $D$, number of samples $n$, perturbation $P$, \\
   \-\hspace{1.4cm}performance test $c$, performance threshold $c^*$, decision rule $R$
   \STATE Perturbation level $i=1$
   \REPEAT 
   \FOR{$j=0$ {\bfseries to} $n-1$}
       \STATE Perturb $X$ using $P$ at level $i$ to generate $X_{j}^{i}$
       \STATE Compute $c(X, X_{j}^{i})$
   \ENDFOR
   \STATE $c(X, X^{i})$ = median($c(X, X_{j}^{i})$)
   \IF{$c(X, X^{i}) \geq c^*$}
            \STATE $i = i+1$
   \ENDIF
   \UNTIL{$c(X, X^{i}) < c^*$}
   \STATE $p^* = i-1$
   \FOR{$j=0$ {\bfseries to} $n-1$}
        \STATE Perturb $X$ using $P$ at level $p^*$ to generate $X_{j}^{p^*}$
        \STATE Compute $D_j^{p^*} = D(X, X_{j}^{p^*})$ for distance function $D$
   \ENDFOR
   \STATE Compute $D^*$ by taking a median (mean) over $D_j^{p^*}$
   \STATE Compute $D_{XY} = D(X, Y)$
   \STATE Determine shift using $R(D_{XY}, D^*)$
\end{algorithmic}
\end{algorithm}

kNN recall is an intuitive metric in this context, because it is directly related to performance in search applications.
For each point, it compares the set of nearest neighbors in the reference and perturbed (evaluation) datasets, respectively.  
Averaging the fraction of neighbors from the original dataset preserved in the evaluation dataset over all points yields the overall kNN recall.
Thus our perturbation test relies on neighborhood structure to find distance threshold. (Note that other performance tests can be substituted for kNN recall.)

\begin{figure}[h]
    \centering
    \includegraphics[width=0.7\textwidth]{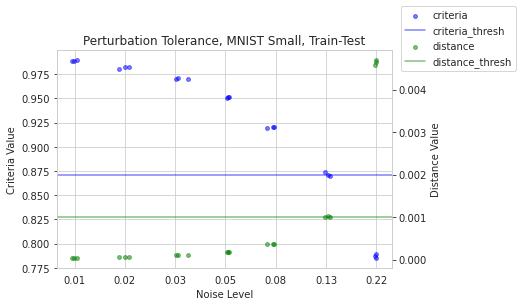}
    \caption{The perturbation test progressively adds noise to a reference dataset, while tracking performance criteria and a distribution distance between reference and noisy versions. When the criteria 
    value dips below the threshold (here $0.80$), the latest valid setting is used to identify a distance threshold (here ~$0.001$) to be used for shift detection.}
    \label{fig:mnist_small_perturbation_sensitivity}
\end{figure}

Figure~\ref{fig:mnist_small_perturbation_sensitivity} demonstrates the behavior of the perturbation test, and shows the interaction among noise level, criteria value, and distribution distance. Adding more noise causes the performance criteria value (e.g. kNN recall) to decrease, while causing distribution distance between reference and noisy datasets to increase.

One limitation of the perturbation test is that each subsample must be representative, requiring either a sophisticated optimal subsampling technique, or the ability to compute over a large percentage of the dataset. Because optimal subsampling is not the focus of this work, for this test only, we use only smaller datasets ($n < 10,000$). 

\textbf{Subsample Shift Test.} This test (Algorithm~\ref{alg:subsample}) arrives at the distance threshold by gathering several subsamples of the reference set to estimate the intra-set variation. Specifically, we compute pairwise distances between reference subsamples to estimate the range of  distances to be considered ``in-distribution". To compare with a candidate dataset, we draw subsamples from each and compute pairwise distances.  If a statistical test can reject the null hypothesis that two arrays of distances, one from pairs of reference subsamples, and one from reference and candidate subsample pairs, come from the same distribution, we determine that a shift has occurred.  

Sample size selection is important, as it affects the consistency and runtime of the test. Larger sample sizes are more representative and lead to more consistent estimates of distribution distance, but increase computation time.

\begin{algorithm}[tb!]
    \caption{Subsample Shift Test}
    \label{alg:subsample}
\begin{algorithmic}
    \STATE {\bfseries Input:} Embeddings $X$, $Y$, distance metric $D$, subsample size $m$, number of samples $n$, \\
    \-\hspace{1.4cm}decision rule $R$
    \FOR{$i=0$ {\bfseries to} $n-1$}
        \STATE Generate sample $x_1^i$ of size $m$ from $X$
        \STATE Generate sample $x_2^i$ of size $m$ from $X$
        \STATE $D_{xx}^i = D(x_1^i, x_2^i)$
    \ENDFOR
    \FOR{$i=0$ {\bfseries to} $n-1$}
        \STATE Generate sample $x^i$ of size $m$ from $X$
        \STATE Generate sample $y^i$ of size $m$ from $Y$
        \STATE $D_{xy}^i = D(x^i, y^i)$
    \ENDFOR
    \STATE Report shift according to a decision rule $R(D_{xx}, D_{xy})$, \\
        for example, if the null hypothesis that $D_{xx}$ and $D_{xy}$ values were generated from the same distribution can be rejected with statistical significance
\end{algorithmic}
\end{algorithm}

\subsection{Distance Metrics}

A successful application of our algorithm depends in a large part on the choice of a distance metric that works well in a high-dimensional space. 
We used three distance metrics in our experiments:

\textbf{Energy Distance} is a nonparametric distribution distance metric \cite{rizzo2016energy}, and is an instance of Maximum Mean Discrepancy with kernel equal to the negative Euclidean norm between elements.
The function requires no assumptions or explicit functional form, and can be easily evaluated over collections of high-dimensional data, such as two sets of embeddings, $X$ and $Y$.

\begin{figure*}
    \hspace*{-1.9cm}
    \centering
    \includegraphics[width=1.2\textwidth]{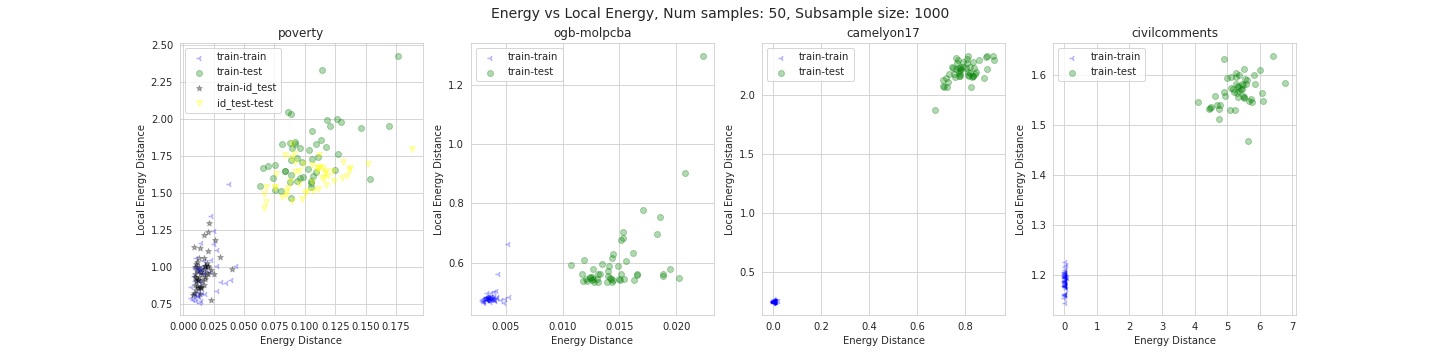}
    \caption{Local energy distance approximates the energy distance metric, and the two are positively correlated when applied to datasets with distribution shift. Both metrics can easily separate train-train from train-test distances, based on the large horizontal and vertical gaps between clusters, indicating their utility for these datasets.}
    \label{fig:local_energy_wilds}
\end{figure*}

\textbf{Local Energy Distance} is a modified form of energy distance that restricts distance computations to each point's closest $k$ neighbors. For example, for sets $X$ and $Y$, and $k$-neighborhood $\mathcal{N}_k(x)$, the squared local energy distance is defined as:
\begin{align}
    E^2(X, Y) = & \mbox{E}_{x \in X, y \in \mathcal{N}_k(x)}[||x - y||] \\ \nonumber
    &+ \mbox{E}_{y \in Y, x \in \mathcal{N}_k(y)}[||y - x||] \\ \nonumber
    &- \mbox{E}_{x, x'}[||x - x'||] - \mbox{E}_{y, y'}[||y - y'||]
\end{align}
Note that as the number of neighbors approaches the dataset size, this converges to the original Energy Distance. In all applications of this function in this work, we use $k=5$. A comparison to energy distance appears in Fig.~\ref{fig:local_energy_wilds}.

\textbf{Sliced Wasserstein Distance on Persistence Diagrams.} 
Persistent homology is a method for approximating topological features of a latent space at different spatial resolutions from a finite set of points in this space.
By considering a ball with a radius value $\mathit{r}$ on each point, a 
simplex is constructed if the balls of any two vertices of the simplex intersect. 
The set of all possible simplices with different dimensions forms a simplicial complex, called Vietoris–Rips complex \cite{Cohen-Steiner} of the 
set of points
with parameter $\mathit{r}$. 
As the radius value  increases from zero, a nested sequence of the simplicial complexes is produced.
It captures features based on connected components, loops and trapped voids.
The extracted topological features are recorded in a topological summary called a persistence diagram, 
a multiset of points in $\mathbb{R}^{2}$. 
A detailed description is available in \cite{Cohen-Steiner}.

One can compute a distance measure between two persistence diagrams using the sliced Wasserstein distance, as defined in \cite{pmlr-v70-carriere17a}. In this work, we compute this measure using the persim implementation within the Scikit-TDA Python package \cite{scikittda2019}.

For the remainder of this work, sliced Wasserstein distance on persistence diagrams will be referred to as SWP.

\section{Experiment Setup}
\label{sec:setup}

\subsection{Datasets and Representation Learning Models}

For the first set of experiments, we use embeddings generated by our own model trained on the MNIST dataset \cite{deng2012mnist}.  The model (detailed further in Appendix~\ref{sec:appendix_experiment_setup}) contains two convolution layers with ReLU activations and max pooling, one fully-connected layer of size $128$, and one classification layer of size $10$; and is trained with cross entropy loss, Adam optimizer, and learning rate $0.001$ for $10$ epochs. Computations are run on a machine with Intel Xeon CPU, 2.2GHz, 48 cores, and 64GB RAM.

Both reference embeddings and candidate embeddings are generated by the model from original reference and candidate datasets.
In some cases, we scale-down such datasets, while preserving desired class distributions. We call shifts evaluated in these experiments \textit{synthetic shifts}, since we control the type and level of distortion in their creation.

For the second set of experiments, we use real-world datasets from the WILDS repository \cite{koh-a}: Camelyon17 \cite{wilds-camelyon17}, CivilComments \cite{wilds-civilcomments}, OGB-MolPCBA \cite{wilds-ogb}, and PovertyMap \cite{wilds-poverty}. These datasets cover unique modalities: biomedical lymph node tissue images, online comment text, molecular graphs, and satellite images, respectively. They also cover subpopulation and domain generalization shift types. See Table~\ref{tab:wilds_datasets} for dataset and model details. For each dataset, we train the model with its original training set and then generate embeddings for 
each specified split.

\begin{table}
\caption{WILDS datasets}
\label{tab:wilds_datasets}
\begin{center}
\begin{small}
\begin{tabular}{lcccc}
\toprule
Name & Shift & Modality & Model & Dim \\
\midrule
Camelyon17 & D & Image & DenseNet121 & 1024 \\
CivilComments & S & Text & DistilBERT & 768\\
OGB-MolPCBA & D & Graph & GIN & 300\\
Poverty & S, D & Image & ResNet18 & 512\\
\bottomrule
\end{tabular}
\end{small}
\vskip 0.05in
D - Domain, S - Subpopulation
\end{center}
\begin{center}
\end{center}
\end{table}

\subsection{Distribution Shifts}
\label{sec:perturbations}

In the experiments on MNIST, we generate synthetic shifts to test our algorithm:
\begin{itemize}
    \item \textbf{Subpopulation Shift} - sample with class distribution different from that of training data, while preserving all classes.  In our experiments, we randomly divide the ten labels into two groups. In the train set, the first group of labels is subset to 10\%, and in the test set, the second group of labels is subset to 10\%. 
    \item \textbf{Domain Shift} - sample different subsets of classes in training and test sets %
    -- a special case of subpopulation shift, with subsampling  down to 0\%, yielding a completely disjoint set of classes in train and test.
\end{itemize}

In the ablation in Fig.~\ref{fig:ablation_shift_magnitude}, we sample from Dirichlet distributions to obtain a wider range of class distributions, and therefore a wider range of shift magnitude.

\subsection{Terminology}

In the tabular results for the MNIST dataset (Table \ref{tab:mnist_synthetic}), ``Shift Type" denotes the synthetic shift applied.
``\underline{Baseline}" applies the shift test to identical train splits to validate the ability to get a true negative result. ``\underline{None}" uses the original train and test splits -- each has a roughly uniform distribution of labels. ``\underline{Subpop}" uses Subpopulation shift, and ``\underline{Domain}" uses Domain shift as defined above.

Tabular results in Tables~\ref{tab:mnist_synthetic},~\ref{tab:real-world} refer to the following terms:\\
\vskip -0.2in
\textbf{Shift Test} refers to the combination of decision algorithm and distance function. We use two tests: Perturbation Shift Test and Subsample Shift Test; and three distance functions: Energy, Local Energy, and Sliced Wasserstein on persistence diagrams.\\
\vskip -0.2in
\textbf{Time} denotes the average runtime in seconds of a single run. Note that subsample shift test contains multiple runs per test but perturbation contains a single run.\\
\vskip -0.2in
\textbf{Metric Ranges} are the $5^{th}$ and $95^{th}$ percentiles of sample measurements of $D_{xx}$ and $D_{xy}$. Here, $D_{xx}$ represents the distance between different subsamples of reference data and $D_{xy}$ represents the distance between subsamples from reference and candidate data.
\\
\vskip -0.2in
\textbf{P-Val Range} is the $5^{th}$ and $95^{th}$ percentiles of p-values from a statistical test with null hypothesis that measurements of $D_{xx}$ and $D_{xy}$ come from the same distribution.\\
\vskip -0.2in
\textbf{Fit Score} is the proportion of subsample runs whose decision matched the majority decision. The fit score represents the confidence of the shift decision.\\
\vskip -0.2in
\textbf{Shift Decision} is a binary output which detects if a distribution shift has occurred for that split pair.\\
\vskip -0.2in
\textbf{Model Performance} is the 
embedding model performance on the task for which it was trained. 
MNIST experiments train and evaluate using splits defined by synthetic shifts. WILDS models are trained on their original training set and evaluated on the set specified in each table row.

\section{Results}
\label{sec:results}
In the following section, we describe the results of applying the Perturbation Shift Test and Subsample Shift Test to MNIST datasets with synthetic shifts, and to WILDS datasets with natural shifts. In all settings, model performance (on the task for which it is trained), is also reported. As validation, we see that in general, as the gap in model performance increases, we are more likely to detect a shift.

For each Perturbation Shift Test, embeddings are normalized, noise perturbations use a log-scaled grid of size $10$ from $0.01$ to $1.0$, and $3$ samples are used to estimate criteria and distance values for that perturbation level. For each Subsample Shift Test, embeddings are normalized, and $20$ runs each using $15$ subsamples of size $1000$ are used.

\subsection{MNIST}
\label{sec:synthetic}

\begin{figure*}[hbt!]
\subfloat[
    \label{fig:ablation_mnist_small_full_range}
    As distribution distance between embedding sets increases, detection sensitivity increases, for a range of sample sizes.
    ]{{
    \includegraphics[width=0.29\textwidth]{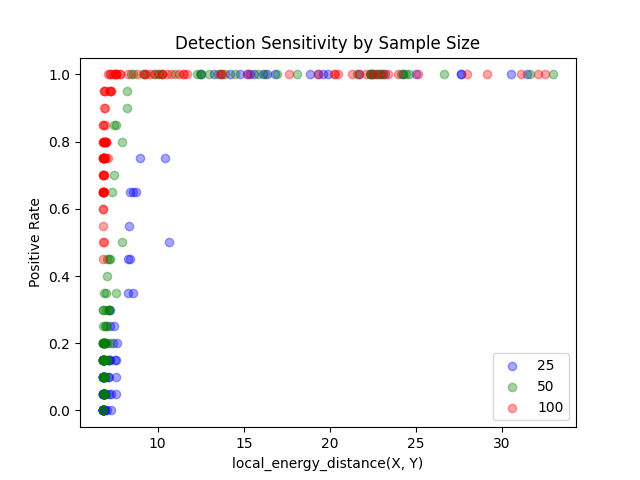}
    }}
\qquad
\subfloat[
    \label{fig:ablation_mnist_small_low_range}
    The positive relationship between distribution distance and detection sensitivity is most visible in the range of small shifts. Larger samples yield high detection sensitivity for even the smallest shifts.]
    {{
    \includegraphics[width=0.29\textwidth]{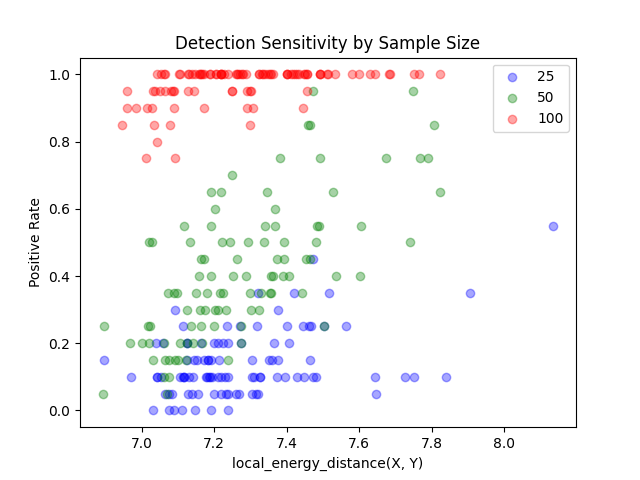}
    }}
\qquad
\subfloat[
    \label{fig:ablation_mnist_small_distances}
    As label distribution distance increases, model accuracy declines and detection sensitivity increases.]
    {{
    \includegraphics[width=0.29\textwidth]{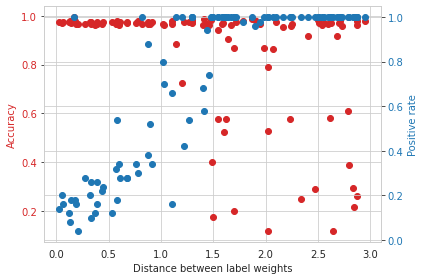}
    }}
\caption{Detection sensitivity (test positive rate) on scaled-down MNIST dataset (n=7000). One hundred runs of each sample size are plotted, for a range of distribution shifts, where class proportions are sampled with Dirichlet distribution.}
\label{fig:ablation_shift_magnitude}
\end{figure*}

In Table~\ref{tab:mnist_synthetic}, both shift tests successfully return negative results (no shift detected) when tested with identical distributions, and return positive results (shift detected) when tested with dataset pairs representing shifts of varying sizes. Most tests report high Fit Scores, indicating that subsample sizes are large enough for unanimous or nearly unanimous agreement across runs. %
P-value ranges from t-tests at significance level $0.05$ align with expected shift decisions.%
Across runs, shift tests with energy and SWP take $4x$ longer and $29x$ longer, respectively, when compared to local energy.

Results reveal that energy behaves differently compared to local energy and SWP.
Interestingly, the energy distance in case ``S-E" for shift type ``None" reports no shift between original train and test splits, while the other two metrics detect shift in small differences between these splits.
Similarly, based on the metric ranges of $D_{xx}$ and $D_{xy}$, energy is less sensitive to domain shifts than to subpopulation shifts.

Figures~\ref{fig:ablation_mnist_small_full_range} and \ref{fig:ablation_mnist_small_low_range} show an ablation on sample size in the Subsample Shift Test. Figure~\ref{fig:ablation_mnist_small_full_range} shows that as sample size increases, the detection sensitivity of the shift test, or ``positive rate", increases. Figure~\ref{fig:ablation_mnist_small_low_range} shows that for small shifts and for each sample size, a positive correlation exists between distribution distance and positive rate. This confirms that the larger the shift, the easier it is to detect.
Figure~\ref{fig:ablation_mnist_small_distances} shows an ablation on MNIST over different label distributions for reference and candidate data. We see that as the distance between label distributions increase, the positive rate increase and the accuracy drops displaying the inverse correlation between model performance and positive rate.

\begin{table*}[hbt!]
\caption{Experiments on the MNIST dataset}
\label{tab:mnist_synthetic}
\vskip 0.15in
\begin{center}
\begin{small}
\begin{sc}
\hspace*{-0.75cm}
\begin{tabular}{lccccccc}
\toprule
Shift & Shift & Time & Metric Ranges & P-Val & Fit Score & Shift & Model Perf. \\
Type & Test & (sec) & $[D_{xx}]:[D_{xy}]$ & Range & (Y:N) & Dec. & Ref, Cand, Gap \\
\midrule
Baseline & S-LE & 1.04 & $[2.79 \, \, 2.83] : [2.79 \, \, 2.84]$ & $[0.14 \, \, 0.94]$ & $1.00 \,\, (0:20)$  & No & $0.994 \,\, 0.994 \,\, 0.000$ \\
None & S-LE & 0.89 & $[2.80 \, \, 2.84] : [2.98 \, \, 3.02]$ & $[0.00 \, \, 0.00]$ & $1.00 \,\, (20:0)$  & Yes & $0.994 \,\, 0.984 \,\, 0.010$ \\
Subpop & S-LE & 0.80 & $[1.86 \, \, 1.89] : [5.00 \, \, 5.24]$ & $[0.00 \, \, 0.00]$ & $1.00 \,\, (20:0)$  & Yes & $0.998 \,\, 0.909 \,\, 0.089$ \\
Domain & S-LE & 0.80 & $[1.51 \, \, 1.54] : [7.17 \, \, 7.38]$ & $[0.00 \, \, 0.00]$ & $1.00 \,\, (20:0)$  & Yes & $0.999 \,\, 0.000 \,\, 0.999$ \\
\midrule
Baseline & S-E & 4.73 & $[0.03 \, \, 0.03] : [0.03 \, \, 0.03]$ & $[0.06 \, \, 0.95]$ & $0.95 \,\, (1:19)$  & No & $0.994 \,\, 0.994 \,\, 0.000$ \\
None & S-E & 4.57 & $[0.03 \, \, 0.03] : [0.03 \, \, 0.04]$ & $[0.01 \, \, 0.92]$ & $0.80 \,\, (4:16)$  & No & $0.994 \,\, 0.984 \,\, 0.010$ \\
Subpop & S-E & 4.52 & $[0.02 \, \, 0.03] : [1.94 \, \, 2.02]$ & $[0.00 \, \, 0.00]$ & $1.00 \,\, (20:0)$  & Yes & $0.998 \,\, 0.909 \,\, 0.089$ \\
Domain & S-E & 4.45 & $[0.02 \, \, 0.03] : [1.36 \, \, 1.43]$ & $[0.00 \, \, 0.00]$ & $1.00 \,\, (20:0)$  & Yes & $0.999 \,\, 0.000 \,\, 0.999$ \\
\midrule
Baseline & S-SWP & 29.02 & $[71.27 \, \, 93.41] : [75.76 \, \, 94.07]$ & $[0.11 \, \, 0.90]$ & $0.95 \,\, (1:19)$  & No & $0.994 \,\, 0.994 \,\, 0.000$ \\
None & S-SWP & 28.71 & $[73.03 \, \, 96.84] : [137.79 \, \, 165.23]$ & $[0.00 \, \, 0.00]$ & $1.00 \,\, (20:0)$  & Yes & $0.994 \,\, 0.984 \,\, 0.010$ \\
Subpop & S-SWP & 27.60 & $[53.48 \, \, 73.04] : [140.00 \, \, 161.11]$ & $[0.00 \, \, 0.00]$ & $1.00 \,\, (20:0)$  & Yes & $0.998 \,\, 0.909 \,\, 0.089$ \\
Domain & S-SWP & 28.16 & $[46.95 \, \, 57.98] : [361.20 \, \, 388.05]$ & $[0.00 \, \, 0.00]$ & $1.00 \,\, (20:0)$  & Yes & $0.999 \,\, 0.000 \,\, 0.999$ \\

\midrule
Baseline & P & 23.01 & $[0.60] : [0.00]$ & - & -  & No & $0.994 \,\, 0.994 \,\, 0.000$ \\
None & P & 23.36 & $[0.60] : [2.43]$ & - & -  & Yes & $0.994 \,\, 0.984 \,\, 0.010$ \\
Subpop & P & 22.65 & $[0.60] : [4.79]$ & - & - & Yes & $0.998 \,\, 0.909 \,\, 0.089$ \\
Domain & P & 23.30 & $[0.62] : [6.65]$ & - & - & Yes & $0.999 \,\, 0.000 \,\, 0.999$ \\
\bottomrule
\end{tabular}
\end{sc}
\end{small}
\end{center}

Results on MNIST datasets are reported for various shift types and shift tests. Abbreviations: Subsample Shift Test with local energy (S-LE), energy (S-E), and sliced Wasserstein on persistence diagrams (S-SWP), respectively, with subsample size $1000$; Perturbation Shift Test with single run on scaled-down MNIST (P).
\end{table*}

\subsection{WILDS}
\label{sec:real-world}

Table \ref{tab:real-world} summarizes our results on WILDS datasets. In this setting, we see similar anticipated results. Both shift tests successfully return negative results when tested with identical distributions, and return positive results when tested with dataset pairs representing shifts. Here too, Fit Scores are high indicating that sample size $1000$ is sufficient to capture these types of shift. An exception occurs with the Poverty dataset, when comparing Train and In-Domain Test splits. 

In this case, the p-values (with $5^{th}$ and $95^{th}$ percentiles of $[0, 0.64]$) span the significance threshold of $0.05$, leading to a $11:9$ mix of shift decisions. Similarly, the metric ranges, $D_{xx} = [3.76, 3.92], D_{xy} = [3.88, 4.00]$, show considerable overlap, justifying the confusion.

This particular test split is designed (by the WILDS authors) to have the same domain as the train split. Despite the shift test's high sensitivity, it returns mixed results in a way that reflects the underlying truth.
Results for runtime show that individual runs with local energy scale with embedding size, and that the Perturbation Shift Test is significantly slower, here taking $\sim228$ seconds when computing for two datasets of size $9797$.
Across runs, shift tests with energy and SWP take $2-5x$ longer and $3-15x$ longer, respectively, when compared to local energy.

\begin{table*}[hbtp!]
\caption{Experiments on WILDS datasets}
\label{tab:real-world}
\vskip 0.15in
\begin{center}
\begin{small}
\begin{sc}
\hspace*{-1.41cm}
\begin{tabular}{lcccccccc}
\toprule
 & Split &  Shift & Time & Metric Ranges & P-Val & Fit Score & Shift  & Model Perf. \\
Data & Pair & Test & (sec) &$[D_{xx}]:[D_{xy}]$ & Range & (Y:N) & Dec. & Ref, Cand, Gap \\
\midrule
\textsc{CAM} & Tr-Tr & S-LE & 16.20 & $[3.19 \, \, 3.24] : [3.18 \, \, 3.25]$ & $[0.09 \, \, 0.89]$ & $1.00 \,\, (0:20)$ & No & $0.965 \,\, 0.965 \,\, 0.000$\\
\textsc{CAM} & Tr-Te & S-LE & 11.91 & $[3.10 \, \, 3.15] : [24.33 \, \, 25.04]$ & $[0.00 \, \, 0.00]$ & $1.00 \,\, (20:0)$  & Yes & $0.965 \,\, 0.481 \,\, 0.484$\\
\midrule
\textsc{CAM} & Tr-Tr & S-E & 35.35 & $[0.05 \, \, 0.08] : [0.05 \, \, 0.08]$ & $[0.10 \, \, 0.96]$ & $1.00 \,\, (0:20)$ & No & $0.965 \,\, 0.965 \,\, 0.000$\\
\textsc{CAM} & Tr-Te & S-E & 31.01 & $[0.05 \, \, 0.09] : [8.82 \, \, 9.33]$ & $[0.00 \, \, 0.00]$ & $1.00 \,\, (20:0)$  & Yes & $0.965 \,\, 0.481 \,\, 0.484$\\
\midrule
\textsc{CAM} & Tr-Tr & S-SWP & 39.79 & $[106.65 \, \, 126.10] : [98.33 \, \, 131.72]$ & $[0.15 \, \, 0.96]$ & $0.95 \,\, (1:19)$  & No & $0.965 \,\, 0.965 \,\, 0.000$\\
\textsc{CAM} & Tr-Te & S-SWP & 35.88 & $[92.93 \, \, 121.90] : [249.67 \, \, 298.48]$ & $[0.00 \, \, 0.00]$ & $1.00 \,\, (20:0)$  & Yes & $0.965 \,\, 0.481 \,\, 0.484$\\
\addlinespace
\addlinespace
\midrule
\midrule
\addlinespace
\textsc{CIV} & Tr-Tr & S-LE & 11.27 & $[5.88 \, \, 5.98] : [5.89 \, \, 5.96]$ & $[0.06 \, \, 0.94]$ & $1.00 \,\, (0:20)$  & No & $0.979 \,\, 0.979 \,\, 0.000$\\
\textsc{CIV} & Tr-Te & S-LE & 9.40 & $[5.88 \, \, 5.95] : [7.36 \, \, 7.48]$ & $[0.00 \, \, 0.00]$ & $1.00 \,\, (20:0)$  & Yes & $0.979 \,\, 0.900 \,\, 0.079$\\
\midrule
\textsc{CIV} & Tr-Tr & S-E & 23.70 & $[0.05 \, \, 0.07] : [0.05 \, \, 0.07]$ & $[0.12 \, \, 0.96]$ & $1.00 \,\, (0:20)$  & No & $0.979 \,\, 0.979 \,\, 0.000$\\
\textsc{CIV} & Tr-Te & S-E & 21.98 & $[0.05 \, \, 0.07] : [8.95 \, \, 9.88]$ & $[0.00 \, \, 0.00]$ & $1.00 \,\, (20:0)$  & Yes & $0.979 \,\, 0.900 \,\, 0.079$\\
\midrule
\textsc{CIV} & Tr-Tr & S-SWP & 36.13 & $[187.81 \, \, 241.98] : [189.56 \, \, 243.23]$ & $[0.05 \, \, 0.92]$ & $0.90 \,\, (2:18)$  & No & $0.979 \,\, 0.979 \,\, 0.000$\\
\textsc{CIV} & Tr-Te & S-SWP & 34.53 & $[172.33 \, \, 233.34] : [577.12 \, \, 660.89]$ & $[0.00 \, \, 0.00]$ & $1.00 \,\, (20:0)$  & Yes & $0.979 \,\, 0.900 \,\, 0.079$\\
\addlinespace
\addlinespace
\midrule
\midrule
\addlinespace
\textsc{OGB} & Tr-Tr & S-LE & 6.05 & $[3.83 \, \, 3.88] : [3.83 \, \, 3.88]$ & $[0.06 \, \, 0.86]$ & $0.95 \,\, (1:19)$  & No & $0.474 \,\, 0.474 \,\, 0.000$\\
\textsc{OGB} & Tr-Te & S-LE & 3.89 & $[3.30 \, \, 3.34] : [3.77 \, \, 3.86]$ & $[0.00 \, \, 0.00]$ & $1.00 \,\, (20:0)$  & Yes & $0.474 \,\, 0.272 \,\, 0.202$\\
\midrule
\textsc{OGB} & Tr-Tr & S-E & 10.83 & $[0.03 \, \, 0.03] : [0.03 \, \, 0.03]$ & $[0.07 \, \, 0.88]$ & $1.00 \,\, (0:20)$  & No & $0.474 \,\, 0.474 \,\, 0.000$\\
\textsc{OGB} & Tr-Te & S-E & 9.05 & $[0.03 \, \, 0.03] : [0.08 \, \, 0.10]$ & $[0.00 \, \, 0.00]$ & $1.00 \,\, (20:0)$  & Yes & $0.474 \,\, 0.272 \,\, 0.202$\\
\midrule
\textsc{OGB} & Tr-Tr & S-SWP & 34.75 & $[98.57 \, \, 121.49] : [101.11 \, \, 143.15]$ & $[0.10 \, \, 0.98]$ & $1.00 \,\, (0:20)$  & No & $0.474 \,\, 0.474 \,\, 0.000$\\
\textsc{OGB} & Tr-Te & S-SWP & 32.59 & $[86.73 \, \, 109.42] : [317.06 \, \, 515.51]$ & $[0.00 \, \, 0.03]$ & $1.00 \,\, (20:0)$  & Yes & $0.474 \,\, 0.272 \,\, 0.202$\\
\addlinespace
\addlinespace
\midrule
\midrule
\addlinespace
\textsc{POV} & Tr-Tr & S-LE & 2.36 & $[3.77 \, \, 3.90] : [3.78 \, \, 3.93]$ & $[0.22 \, \, 0.94]$ & $1.00 \,\, (0:20)$  & No & $0.908 \,\, 0.908 \,\, 0.000$\\
\textsc{POV} & Tr-Te$^{i}$ & S-LE & 2.25 & $[3.76 \, \, 3.92] : [3.88 \, \, 4.00]$ & $[0.00 \, \, 0.64]$ & $0.55 \,\, (11:9)$  & Yes & $0.908 \,\, 0.867 \,\, 0.041$\\
\textsc{POV} & Tr-Te & S-LE & 2.22 & $[3.67 \, \, 3.75] : [7.25 \, \, 7.40]$ & $[0.00 \, \, 0.00]$ & $1.00 \,\, (20:0)$  & Yes & $0.908 \,\, 0.845 \,\, 0.063$\\
\textsc{POV} & Te$^{i}$-Te & S-LE & 2.12 & $[0.00 \, \, 0.00] : [6.61 \, \, 6.76]$ & $[0.00 \, \, 0.00]$ & $1.00 \,\, (20:0)$  & Yes & $0.867 \,\, 0.845 \,\, 0.022$\\
\midrule
\textsc{POV} & Tr-Tr & S-E & 11.90 & $[0.04 \, \, 0.05] : [0.04 \, \, 0.06]$ & $[0.09 \, \, 0.95]$ & $1.00 \,\, (0:20)$  & No & $0.908 \,\, 0.908 \,\, 0.000$\\
\textsc{POV} & Tr-Te$^{i}$ & S-E & 11.89 & $[0.04 \, \, 0.05] : [0.04 \, \, 0.05]$ & $[0.02 \, \, 0.66]$ & $0.80 \,\, (4:16)$  & No & $0.908 \,\, 0.867 \,\, 0.041$\\
\textsc{POV} & Tr-Te & S-E & 11.93 & $[0.04 \, \, 0.05] : [0.38 \, \, 0.43]$ & $[0.00 \, \, 0.00]$ & $1.00 \,\, (20:0)$  & Yes & $0.908 \,\, 0.845 \,\, 0.063$\\
\textsc{POV} & Te$^{i}$-Te & S-E & 11.92 & $[0.00 \, \, 0.00] : [0.35 \, \, 0.38]$ & $[0.00 \, \, 0.00]$ & $1.00 \,\, (20:0)$  & Yes & $0.867 \,\, 0.845 \,\, 0.022$\\
\midrule
\textsc{POV} & Tr-Tr & S-SWP & 31.96 & $[161.52 \, \, 208.24] : [160.98 \, \, 208.52]$ & $[0.10 \, \, 0.83]$ & $1.00 \,\, (0:20)$  & No & $0.908 \,\, 0.908 \,\, 0.000$\\
\textsc{POV} & Tr-Te$^{i}$ & S-SWP & 31.98 & $[162.90 \, \, 202.68] : [236.59 \, \, 292.11]$ & $[0.00 \, \, 0.05]$ & $0.90 \,\, (18:2)$  & Yes & $0.908 \,\, 0.867 \,\, 0.041$\\
\textsc{POV} & Tr-Te & S-SWP & 30.83 & $[152.27 \, \, 183.88] : [465.03 \, \, 518.05]$ & $[0.00 \, \, 0.00]$ & $1.00 \,\, (20:0)$  & Yes & $0.908 \,\, 0.845 \,\, 0.063$\\
\textsc{POV} & Te$^{i}$-Te & S-SWP & 30.73 & $[0.00 \, \, 0.00] : [526.06 \, \, 569.28]$ & $[0.00 \, \, 0.00]$ & $1.00 \,\, (20:0)$  & Yes & $0.867 \,\, 0.845 \,\, 0.022$\\
\midrule
\textsc{POV} & Tr-Tr & P & 227.89 & $[0.72]:[0.00]$     & - & -  & No & $0.908 \,\, 0.908 \,\, 0.000$\\
\textsc{POV} & Tr-Te$^{i}$ & P & 259.41 & $[0.71]:[3.19]$     & - & -  & Yes & $0.908 \,\, 0.867 \,\, 0.041$\\
\textsc{POV} & Tr-Te & P & 222.26 & $[0.72]:[6.34]$     & - & -  & Yes & $0.908 \,\, 0.845 \,\, 0.063$\\
\textsc{POV} & Te$^{i}$-Te & P & 131.17 & $[1.22]:[6.40]$     & - & -  & Yes & $0.867 \,\, 0.845 \,\, 0.022$\\
\bottomrule
\end{tabular}
\end{sc}
\end{small}
\end{center}
Results on WILDS datasets (details in Table~\ref{tab:wilds_datasets}) are reported for various split pairs and shift tests. Abbreviations: Train (Tr), Test (Te), In-domain test (Te$^{i}$), Subsample Shift Test with local energy (S-LE), Subsample Shift Test with energy (S-E), Subsample Shift Test with sliced Wasserstein on persistence diagrams (S-SWP), and Perturbation Shift Test with single run on Poverty dataset (P). All Subsample Shift Tests use subsample size $1000$. 
\end{table*}

\section{Discussion}
\label{sec:conclusion}

We have proposed a framework for discovering distribution shifts in data, by analyzing the embedding spaces resulting from applying a model to reference and candidate datasets.  Our non-parametric framework is fully general and can be tuned for a specific domain. %
Experiments show that this framework can be successfully applied to both synthetic and real-world shift scenarios.

When we discuss distribution shift detection, we assume that the decision is always binary: a shift has occurred or it did not.  However, in many cases, a shift is a matter of degree.  Data sampling often results in datasets exhibiting small variation.%
We may not want to label such variations as ``shifts".  Shift magnitude that warrants detection is also likely to differ from one domain to another.  Our framework allows for adjusting detection sensitivity by changing subsample size, distance metric, and/or decision threshold. In practice, Fit Score can be used to verify that these settings result in the correct level of sensitivity.

Not surprisingly, test sensitivity increases with larger subsample sizes, as demonstrated in Figure \ref{fig:ablation_mnist_small_low_range}. This sensitivity comes at the expense of computation time. 
Although larger subsamples give us more information on the distribution of pairwise distances, shifts can occur on different scales. 
Because the local energy distance is only exposed to local neighborhoods, it is
oblivious to changes in the diameter of the embedding space corresponding to the tail of the distribution. %
In contrast, the energy distance is exposed to the full dataset, but only in the form of summary statistics,
and may therefore be insensitive to non-uniform local perturbations in the embedding space, as each point has less leverage.
Approaches that take all scales into account, such as the SWP, are flexible enough for all types of shifts, but are often more computationally expensive. An advantage to remember, however, is that computation scales not with dataset size, but with the subsample size required to represent the distribution.
One crucial observation is that all metrics tested were able to obtain accurate results with subsamples of relatively small size.
We suspect that metrics of this type -- which operate on matrices of pairwise distances -- 
will perform well given this framework. Furthermore, since this family of distances has a computational complexity $\geq O(n^2)$, it is computationally pragmatic to use more subsamples, as opposed to increasing the size of each subsample. This suggestion coincides with our empirical findings. 

Finally, we note again the practicality of such an approach, which presents fast and accurate shift detection on modest hardware that is available to most users.

\section{Future Work}
\label{sec:future_work}
Future work includes performing experiments for less-frequently discussed shifts. General density distortion, 
where we preferentially sample from subareas of the space, is one such case.  General density distortion generalizes subpopulation shift -- we consider altering the distribution not just for classes (e.g. digits in MNIST), but also for subareas within these classes, for class boundary areas, and for arbitrary subspaces.

%

%
%
%

%
%

%
%
%


\bibliography{embeddings}

\begin{thebibliography}{10}

\bibitem{DShift-EHR-Bench}
A.~Avati, M.~Seneviratne, E.~Xue, Z.~Xu, B.~Lakshminarayanan, and A.~M. Dai.
\newblock {BEDS}-bench: Behavior of ehr-models under distributional shift–a
  benchmark.
\newblock In {\em Workshop on Distribution Shifts, 35th Conference on Neural
  Information Processing Systems (NeurIPS)}, 2021.

\bibitem{wilds-camelyon17}
P.~Bandi, O.~Geessink, Q.~Manson, M.~Van~Dijk, M.~Balkenhol, M.~Hermsen, B.~E.
  Bejnordi, B.~Lee, K.~Paeng, A.~Zhong, et~al.
\newblock From detection of individual metastases to classification of lymph
  node status at the patient level: the {CAMELYON17} challenge.
\newblock {\em IEEE Transactions on Medical Imaging}, 2018.

\bibitem{wilds-civilcomments}
D.~Borkan, L.~Dixon, J.~Sorensen, N.~Thain, and L.~Vasserman.
\newblock Nuanced metrics for measuring unintended bias with real data for text
  classification.
\newblock In {\em Companion Proceedings of The 2019 World Wide Web Conference},
  2019.

\bibitem{DShift-BuresDiv}
A.~J. Brockmeier, C.~C. Claros-Olivares, M.~S. Emigh, and L.~G.~S. Giraldo.
\newblock Identifying the instances associated with distribution shifts using
  the max-sliced bures divergence.
\newblock In {\em Workshop on Distribution Shifts, 35th Conference on Neural
  Information Processing Systems (NeurIPS)}, 2021.

\bibitem{pmlr-v70-carriere17a}
M.~Carri{\`e}re, M.~Cuturi, and S.~Oudot.
\newblock Sliced {W}asserstein kernel for persistence diagrams.
\newblock In D.~Precup and Y.~W. Teh, editors, {\em Proceedings of the 34th
  International Conference on Machine Learning}, volume~70 of {\em Proceedings
  of Machine Learning Research}, pages 664--673. PMLR, 06--11 Aug 2017.

\bibitem{Cohen-Steiner}
D.~Cohen-Steiner, H.~Edelsbrunner, and J.~Harer.
\newblock Stability of persistence diagrams.
\newblock {\em Discrete \& Computational Geometry}, 37:103--120, 2006.

\bibitem{deng2012mnist}
L.~Deng.
\newblock The {MNIST} database of handwritten digit images for machine learning
  research.
\newblock {\em IEEE Signal Processing Magazine}, 29(6):141--142, 2012.

\bibitem{DShift-Graph-Bench}
M.~Ding, K.~Kong, J.~Chen, J.~Kirchenbauer, M.~Goldblum, D.~Wipf, F.~Huang, and
  T.~Goldstein.
\newblock A closer look at distribution shifts and out-of-distribution
  generalization on graphs.
\newblock In {\em Workshop on Distribution Shifts, 35th Conference on Neural
  Information Processing Systems (NeurIPS)}, 2021.

\bibitem{geirhos2020a}
R.~Geirhos, J.~Jacobsen, C.~Michaelis, R.~Zemel, W.~Brendel, M.~Bethge, and
  F.~Wichmann.
\newblock Shortcut learning in deep neural networks.
\newblock {\em CoRR}, 2020.

\bibitem{DShift-OOD-Detection}
E.~D.~C. Gomes, F.~Alberge, P.~Duhamel, and P.~Piantanida.
\newblock \textsc{Igeood}: An information geometry approach to
  out-of-distribution detection.
\newblock In {\em Workshop on Distribution Shifts, 35th Conference on Neural
  Information Processing Systems (NeurIPS)}, 2021.

\bibitem{hendrycks2019a}
D.~Hendrycks and T.~Dietterich.
\newblock Benchmarking neural network robustness to common corruptions and
  perturbations.
\newblock In {\em International Conference on Learning Representations (ICLR)},
  2019.

\bibitem{wilds-ogb}
W.~Hu, M.~Fey, M.~Zitnik, Y.~Dong, H.~Ren, B.~Liu, M.~Catasta, and J.~Leskovec.
\newblock Open graph benchmark: Datasets for machine learning on graphs.
\newblock In {\em Advances in Neural Information Processing Systems (NeurIPS)},
  2020.

\bibitem{koh-a}
P.~W. Koh, S.~Sagawa, H.~Marklund, S.~M. Xie, M.~Zhang, A.~Balsubramani, W.~Hu,
  M.~Yasunaga, R.~L. Phillips, I.~Gao, T.~Lee, E.~David, I.~Stavness, W.~Guo,
  B.~A. Earnshaw, I.~S. Haque, S.~Beery, J.~Leskovec, A.~Kundaje, E.~Pierson,
  S.~Levine, C.~Finn, and P.~Liang.
\newblock Wilds: A benchmark of in-the-wild distribution shifts.
\newblock In {\em International Conference on Learning Representations (ICLR)},
  2021.

\bibitem{liu2020learning}
F.~Liu, W.~Xu, J.~Lu, G.~Zhang, A.~Gretton, and D.~J. Sutherland.
\newblock Learning deep kernels for non-parametric two-sample tests.
\newblock In {\em International Conference on Machine Learning}, pages
  6316--6326. PMLR, 2020.

\bibitem{DShift-AttribAlignment}
M.~L. Olson, S.~Liu, R.~Anirudh, J.~J. Thiagarajan, W.-K. Wong, and P.-T.
  Bremer.
\newblock Unsupervised attribute alignment for characterizing distribution
  shift.
\newblock In {\em Workshop on Distribution Shifts, 35th Conference on Neural
  Information Processing Systems (NeurIPS)}, 2021.

\bibitem{qui2009a}
J.~Quiñonero-Candela, M.~Sugiyama, A.~Schwaighofer, and N.~Lawrence.
\newblock {\em Dataset Shift in Machine Learning}.
\newblock The MIT Press, 2009.

\bibitem{rizzo2016energy}
M.~L. Rizzo and G.~J. Sz{\'e}kely.
\newblock Energy distance.
\newblock {\em Wiley Interdisciplinary Reviews: Computational Statistics},
  8(1):27--38, 2016.

\bibitem{scikittda2019}
N.~Saul and C.~Tralie.
\newblock Scikit-tda: Topological data analysis for python, 2019.

\bibitem{schrab2021mmd}
A.~Schrab, I.~Kim, M.~Albert, B.~Laurent, B.~Guedj, and A.~Gretton.
\newblock {MMD} aggregated two-sample test.
\newblock {\em arXiv preprint arXiv:2110.15073}, 2021.

\bibitem{vu-a}
X.-S. Vu, T.~Vu, S.~N. Tran, and L.~Jiang.
\newblock {ETNLP}: a visual-aided systematic approach to select pre-trained
  embeddings for a downstream task.
\newblock In {\em Proceedings of Recent Advances in Natural Language
  Processing}, 2019.

\bibitem{wilds-poverty}
C.~Yeh, A.~Perez, A.~Driscoll, G.~Azzari, Z.~Tang, D.~Lobell, S.~Ermon, and
  M.~Burke.
\newblock Using publicly available satellite imagery and deep learning to
  understand economic well-being in africa.
\newblock {\em Nature Communications}, 2020.

\end{thebibliography}
\bibliographystyle{abbrv}

\newpage
\appendix
\onecolumn
\section{Experiment Setup}
\label{sec:appendix_experiment_setup}

The PyTorch model used in synthetic experiments with MNIST image data has the following architecture:

\begin{verbatim}
CNN(
  (conv1): Sequential(
    (0): Conv2d(1, 16, kernel_size=(5, 5), stride=(1, 1), padding=(2, 2))
    (1): ReLU()
    (2): MaxPool2d(kernel_size=2, stride=2, padding=0, dilation=1, ceil_mode=False)
  )
  (conv2): Sequential(
    (0): Conv2d(16, 32, kernel_size=(5, 5), stride=(1, 1), padding=(2, 2))
    (1): ReLU()
    (2): MaxPool2d(kernel_size=2, stride=2, padding=0, dilation=1, ceil_mode=False)
  )
  (fc1): Linear(in_features=1568, out_features=128, bias=True)
  (out): Linear(in_features=128, out_features=10, bias=True)
)
\end{verbatim}

\section{Ablations}
\label{sec:appendix_ablations}

\begin{figure*}[ht]
\subfloat[
    \label{fig:appendix_ablation_mnist_small_full_range}
    As distribution distance between embedding sets increases, detection sensitivity increases, for a range of sample sizes.
    ]{{
    \includegraphics[width=0.3\textwidth]{figures/ablation_plot_local_energy_distance_rep100_n20_ss25-50-100.alpharange.png}
    }}
\qquad
\subfloat[
    \label{fig:appendix_ablation_mnist_small_low_range}
    The positive relationship between distribution distance and detection sensitivity is most visible in the range of small shifts. Larger samples yield high detection sensitivity for even the smallest shifts.]
    {{
    \includegraphics[width=0.3\textwidth]{figures/ablation_plot_local_energy_distance_rep100_n20_ss25-50-100.png}
    }}
\qquad
\subfloat[
    \label{fig:appendix_ablation_mnist_small_distances}
    As label distribution distance increases, model accuracy declines and detection sensitivity increases.]
    {{
    \includegraphics[width=0.3\textwidth]{figures/ablation_mnist_small_l2_acc_posrate.png}
    }}
\qquad
\subfloat[
    \label{fig:appendix_ablation_mnist_small_full_range_e}
    Energy distance yields similar behavior. Compare to Fig.~\ref{fig:appendix_ablation_mnist_small_full_range}.
    ]{{
    \includegraphics[width=0.3\textwidth]{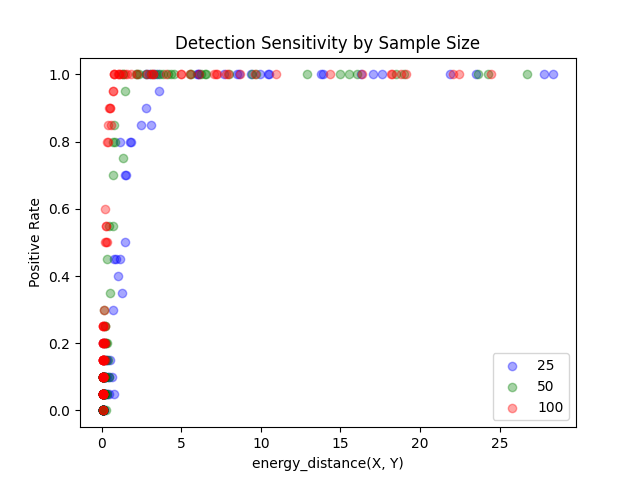} }}
\qquad
\subfloat[
    \label{fig:appendix_ablation_mnist_small_low_range_e}
    Energy distance yields similar behavior. Compare to Fig.~\ref{fig:appendix_ablation_mnist_small_low_range}.
    ]{{
    \includegraphics[width=0.3\textwidth]{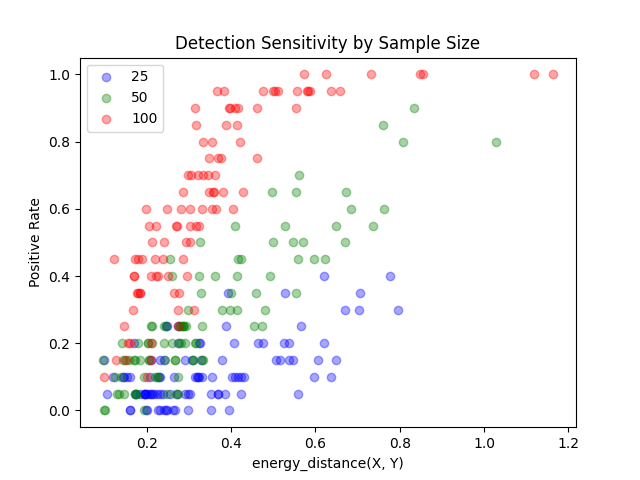}
    }}%
\caption{Detection sensitivity on reduced MNIST dataset (n=7000). One hundred runs per setting are plotted, representing a range of distribution shifts. Note in Fig.~\ref{fig:appendix_ablation_mnist_small_low_range} and Fig.~\ref{fig:appendix_ablation_mnist_small_low_range_e} that for the smallest shifts, local energy reports higher sensitivity ($\sim0.7$) compared to energy ($\sim0.4$).}
\label{fig:example-appendix}%
\end{figure*}

Full results of the ablation, with both energy and local energy, are shown in Fig.~\ref{fig:appendix_ablation_mnist_small_full_range}-\ref{fig:appendix_ablation_mnist_small_low_range_e}.

\end{document}